\documentclass[conference]{IEEEtran}
\IEEEoverridecommandlockouts
\usepackage{cite}
\usepackage{amsmath,amssymb,amsfonts}
\usepackage{algorithmic}
\usepackage{graphicx}
\usepackage{textcomp}
\usepackage{xcolor}
\usepackage{float}
\def\BibTeX{{\rm B\kern-.05em{\sc i\kern-.025em b}\kern-.08em
    T\kern-.1667em\lower.7ex\hbox{E}\kern-.125emX}}

\usepackage{caption}
\usepackage{subcaption}

\DeclareMathOperator{\Real}{Re}
\DeclareMathOperator{\Imag}{Im}

\begin{document}

\title{Socio-cognitive Optimization of Time-delay Control Problems using Evolutionary Metaheuristics
{
\thanks{The research presented in this paper was partially supported by: 
NCN project no: 2020/39/I/ST7/02285, NCN project: no:2019/35/O/ST6/00570, Polish Ministry of Education and Science funds assigned to AGH University of Science and Technology.}
}}

\author{\IEEEauthorblockN{Piotr Kipi\'nski, Hubert Guzowski, Aleksandra Urba\'nczyk, Maciej Smo{\l}ka,\\ Marek Kisiel-Dorohinicki and Aleksander Byrski}
\IEEEauthorblockA{\textit{Institute of Computer Science},
\textit{AGH University of Science and Technology},
Krakow, Poland \\
\{kipinski,guzowski\}@student.agh.edu.pl, \{aurbanczyk,smolka,doroh,olekb\}@agh.edu.pl}
\IEEEauthorblockN{Zuzana Kominkova Oplatkova, Roman Senkerik, Libor Pekar, Radek Matusu, Frantisek Gazdos}
\IEEEauthorblockA{\textit{Faculty of Applied Informatics},
\textit{Tomas Bata University in Zlín},
Czech Republic \\
\{oplatkova,senkerik,pekar,rmatusu,gazdos\}@utb.cz}
}

\maketitle

\begin{abstract}
Metaheuristics are universal optimization algorithms which should be used for solving difficult problems, unsolvable by classic approaches. In this paper we aim at constructing novel socio-cognitive metaheuristic based on castes, and apply several versions of this algorithm to optimization of time-delay system model. Besides giving the background and the details of the proposed algorithms we apply them to optimization of selected variants of the problem and discuss the results.
\end{abstract}


\section{Introduction}
Evolutionary metaheuristics proved to be universal global optimization algorithms. This claim is supported not only by textbooks and experimental research (e.g. \cite{goldberg1995algorytmy,michalewicz}) but also by extensive theoretical works (e.g. \cite{vose1999simple})
showing such algorithms as not only efficient and efficacious algorithms, but also easy-to-understand
nature-inspired algorithms stemming from Darwin and of course Holland works \cite{holland1962outline}.

Following works of Talbi \cite{talbi2009metaheuristics} and considering famous No free lunch theorem
\cite{wolpert1997no}, we are convinced it is beneficial to seek new algorithms which may be applicable
to different new problems better than other ones. For example solving transport problems with Ant Colony Optimization may be as good as with Evolutionary Algorithms, however it is much more natural because of the inherent structure of the ACO (representation in a form of pheromone table is more feasible for transport problems than the genotype-based one). 

Providing we do not forget about seminal work of Sorensen \cite{sorensen}, we can propose novel hybrid algorithms and explore their applicability to different problems. In this paper we focus on time-delay systems, a problem stemming from the area of automatics (cf. e.g. \cite{libor}), applying evolutionary algorithm for solving the task of optimizing its parameters, comparing the results with a recently proposed hybrid evolutionary algorithm, constructed based on psychological inspirations.

Socio-cognitive algorithms \cite{PRACA_BYRSKI_SOCJO_KOGNITIVE} are inspired by the works of Albert Bandura, a famous Canadian/US psychologist, in particular on his theory of social-cognitive learning, assuming that we not only learn from our experiences, but we also observe others. Thus incorporating different methods of getting inspired by parts of populations of evolutionary algorithms (e.g. copying in a different way elements of the solutions of the other parts) lead us to propose different socio-cognitive hybrid algorithms, in this work we focus on an idea of a caste-based algorithm, which may be perceived as a concept related somehow to a parallel evolutionary algorithm \cite{cantu-paz1998survey}, with ,,overlapping'' islands.

In this paper presents, we focus first on a short review of evolutionary and hybrid metaheuristics related to this research, then we present the psychological inspirations leading to socio-cognitive hybrids, then we show the definition of the problem being solved and the idea of the proposed algorithm, finally we show the experimental results and conclude the paper.

\section{Metaheuristics inspired by evolution}

A variety of methods can be employed for solving parametric optimization problems. One of the classic approaches is to traverse downhill the cost functions landscape in iterative steps. This can be achieved by calculating the cost functions local gradient and choosing the next calculation point accordingly. This approach is implemented among others by the conjugate gradient method, BFGS algorithm, and its variant L-BFGS-B \cite{NumericalOptimization}. However, when the function values are uncertain, subject to noise, or otherwise non-smooth, the derivative-free algorithms have to be used. One of the best known among them is the Nelder Mead method which utilizes comparisons of values at vertices of a simplex \cite{NeldMead65}.

Methods based on downhill traversal can yield a good reduction in cost function value using a relatively small amount of evaluations but have limited ability to explore multiple local optimas. Therefore more complex problems that can be highly multimodal require using so-called global optimization methods. Those methods use varied stochastic operations to achieve a better exploration of the functions landscape. Some of the most popular among those methods are swarm and evolutionary algorithms.


Classic evolutionary algorithm is a metaheuristic that mimics the processes of natural evolution \cite{holland1992adaptation}. It operates in a loop on a population of individuals (represented by genotypes) who are subjected to the processes of mutation, crossover reproduction, and selection. In order to reach better performance, this standard version of the algorithm is a constant subject of modifications. It is usually done either through bringing novelty on the base of standard EA parameters or through hybridising EA with different heuristic or metaheuristic algorithms\cite{Talbi2002}.


Some of the state-of-the-art optimization methods utilizing the concept of evolution are variants of differential evolution \cite{DifferentialEvolution} and the CMA evolution strategy \cite{CMA-ES}.


One group of algorithms that are relevant to our research are distributed EAs, such as a model of island EA \cite{sudholt2015parallel}. The idea behind that concept is to divide the population of individuals into subpopulations, let them evolve in parallel and occasionally exchange members. The difference between and island model and caste model incorporated here lays in the exchange of genotype between subpopulations. Between islands occurs a migration of specific individuals who then become members of different island. In our algorithms we decided to use other operators to influence populations. It is either a crossover operator (A) child and a specially designed socio-cognitive mutation operator that allows learning from individuals of different caste (B). Both of them, as well as our third idea of TOPSIS-like mutation (C) are part of a trend of socio-cognitive computing. The idea inspired by the work of Bandura \cite{bandura1986} has already been a successful source of hybrid and modified algorithms of Ant Colony Optimisation \cite{BYRSKI2017397}, Particle Swarm Optimisation \cite{BUGAJSKI2016804} but also for Evolution Strategies \cite{urbanczyk2021socio}. Especially the last one position is worth mentioning, not only because it is also based on an algorithm from an evolutionary family, but because a similar to the TOPSIS, however more primitive mechanism of learning to avoid the worst solutions was incorporated there.

\section{Time-delay optimization problem}
\label{sec:optProblem}
The considered time-delay identification problem \cite{libor} is governed by a model function $G_m: \mathbb{C} \to \mathbb{C}$, namely
\begin{equation}
\label{eq:tdModelFunction}
    G_{m, \mathbf{p}}(s) = 
    \frac{b_0 + b_{0, \tau} e^{-\tau_0 s}}
    {s^3 + a_2 s^2 + a_1 s + a_0 + a_{0, \theta} e^{-\theta s}}
    e^{-\tau s}.
\end{equation}
Parameters of such a model form a 9-dimensional real vector
\begin{equation}
\label{eq:parameters}
    \mathbf{p} = \left[
    b_0, b_{0,\tau},
    \tau_0, \tau,
    a_2, a_1, a_0, a_{0,\theta}, \theta
    \right]. 
\end{equation}
As usual, we assume that some of the parameters are related due to the static gain, i.e.
\begin{equation}
\label{eq:staticGain}
    k = \frac{b_0 + b_{0,\tau}}{a_0 + a_{0,\theta}},
\end{equation}
where the value of $k$ is well known (or estimated).
In our case we used the value $k = 0.0322$. To achieve appropriate properties of solutions (such as stability, feasibility, and minimum-phase conditions) we use the following constraints \cite{TdsStability}:
\begin{equation}
\label{eq:constraints}
\begin{aligned}
    & \tau_0 > 0, \tau > 0, \theta > 0, \\
    & a_2 > 0, a_1 > 0, a_0 + a_{0,\theta} > 0, \\
    & a_2 a_1 > a_0, \\
    & a_2 a_1 > a_0 + a_{0,\theta}, \\
    & \frac{a_{0,\theta}}
    {\sqrt{(a_0 - a_2 \omega^2)^2 + \omega^2 (a_1 - \omega^2)^2}}
    < 1, \quad \forall \omega > 0, \\
    & | b_0 | > | b_{0, \tau} |, \\
    & a_0 \neq 0, a_{0,\theta} \neq 0, b_{0,\tau} \neq 0.
\end{aligned}
\end{equation}
Our main task is to find such parameter values that
\begin{equation}
\label{eq:identificationProblem}
    G_{m, \mathbf{p}} (\mathbf{j} \, \omega_i) =
    A_i + \mathbf{j} \, B_i
\end{equation}
for some $\omega_1, \dots, \omega_n$ and some measured values of $A_1, \dots, A_n$ and $B_1, \dots, B_n$, where $\mathbf{j}$ is the imaginary unit ($\mathbf{j}^2 = -1$).

To solve \eqref{eq:identificationProblem} using optimization methods we reformulate it using the classical least-square approach. It consists in the construction of a cost (or loss) function, which in our case reads
\begin{multline}
    \mathcal{C}(\mathbf{p}) =
    \sum_{i=1}^n
    \left[
    \left( 
    \Real G_{m, \mathbf{p}}(\mathbf{j} \, \omega_i) - A_i
    \right)^2 \right. \\ 
    +
    \left.
    \left( 
    \Imag G_{m, \mathbf{p}}(\mathbf{j} \, \omega_i) - B_i    
    \right)^2
    \right]
\end{multline}
This way we obtain the final version of our main problem, which is to find such parameter values $\mathbf{p}^*$ that
\begin{equation}
    \mathcal{C}(\mathbf{p}^*) =
    \min_{\mathbf{p} \in \mathcal{D}} \mathcal{C}(\mathbf{p}),
\end{equation}
where $\mathcal{D}$ is the set of all $\mathbf{p} \in \mathbb{R}^9$ satisfying \eqref{eq:staticGain} and  \eqref{eq:constraints}.


\section{Caste-based algorithm}
The name of the algorithm comes from the phenomenon of castes -- closed social strata to which affiliation is hereditary \cite{caste}. Castes have existed and exist in different societies, but are especially characteristic of Indian society, where the caste system is perpetuated by the traditional taboos of Hinduism. 

The idea of caste in evolutionary algorithms mimics the caste-divided societies by implementing the division of society according to various criteria. The castes introduce a partial division into the population of individuals. One effect of the introduction of castes is  limiting of the possibility of reproduction to the caste of the individuals, and of course producing the offspring belonging to the same caste. Such an algorithm would only copy most of the ideas of the parallel evolutionary algorithms (and in fact, those two approaches can be compared), however in the case of caste-based algorithms those subpopulation (castes) overlap, as indeed, there exists a possibility of reproduction between the individuals belonging to different castes. 

\subsection{Overlapping caste-based algorithm}
The castes being a result of a breakdown of the initial population are of the same size. There exists a parameterized probability of choosing the parent from another caste. As it was mentioned the membership of the caste is transferred to the next population, and when the individual has parents from two different castes, the membership is decided randomly.

The main steps of the caste-based algorithm are quite similar to the classic evolutionary one \cite{goldberg1995algorytmy}, however of course dedicated steps related to caste-based processing are added or modified:
\begin{enumerate}
    \item The population of individuals is randomly initialized.
    \item Castes are assigned to the individuals.
    \item For each of castes:
    \begin{enumerate}
    \item Selection.
    \item Crossover and mutation, there exists a probability of crossing-over with the individuals from different castes.
    \end{enumerate}
    \item Evaluation is executed for all individuals.
    \item Back to point 3  unless the stopping condition is met.
\end{enumerate}
Caste-based algorithm is implemented in 2 variations -- random caste assignment and elitist caste assignment.  In the former case, all the castes are assigned randomly, while in the latter, castes are assigned according to individual's fitness value. The best individuals are assigned together to the caste no 1, a little bit worse to the caste no 2, and so on.

\subsection{Separated caste-based algorithm}
Separated caste-based algorithm is a modification of the proposed basic algorithm. The main difference is that every caste is developing completely independently, so a certain means of relation among the castes must be introduced, otherwise it would be actually a sequential run of a number of evolutionary algorithms. To introduce certain information sharing, we propose a dedicated  learning operator (fitting in the framework of the social-cognitive learning by Bandura). 

The main idea of the algorithm is that the individuals are learning from higher castes by copying certain genotypes of the individuals. Higher castes have better fitness, and the position of castes is updated throughout the run (so this relation of having higher or lower fitness affects the order of the castes). The algorithm steps are similar to the ones described in the previous section, however the learning operator and the reassignment of the castes are added to the main course of the algorithm.

The whole algorithm looks now as follows:
\begin{enumerate}
    \item The population of individuals is randomly initialized.
    \item Castes are assigned to the individuals.
    \item For each of castes:
    \begin{enumerate}
    \item Selection.
    \item Crossover and mutation, there exists a probability of crossing-over with the individuals from different castes.
    \item For each of the individual:
    \begin{enumerate}
    \item Learning from other individuals (from other castes) with certain probability. 
    If this condition is true, the genes are copied, again  with dedicated, certain probability.
    \end{enumerate}
    \end{enumerate}
    \item Evaluation is executed for all individuals.
    \item Reassigning the order of castes when certain interval is reached.
    \item Back to point 3  unless the stopping condition is met.
\end{enumerate}

\subsection{TOPSIS evolutionary algorithm}
Additional method fitting in the class of socio-cognitive learning methods we used  is TOPSIS.
The Technique for Order of Preference by Similarity to Ideal Solution (TOPSIS) is a multi-criteria decision analysis method, which was originally developed by Ching-Lai Hwang and Yoon in 1981 \cite{hwang-yoon} with further developments by Yoon in 1987,\cite{yoon}and Hwang, Lai and Liu in 1993.\cite{hwang-lai-liu} TOPSIS is based on the concept that the chosen alternative should have the shortest geometric distance from the positive ideal solution (PIS)\cite{assari-mahesh} and the longest geometric distance from the negative ideal solution (NIS) \cite{assari-mahesh}.

We implemented this idea in our algorithm by implementing ,,gravity/anti-gravity mutation'' which is executed on population. Individuals have a chance to be pulled towards the best solution/point and pushed away from the worst solution/point. This point is calculated as a weighted average (4 variants) of X best/worst individuals. 
The probability, strength of influence and the number of the best and worst individuals are determined by dedicated parameters:
\begin{itemize}
    \item p: float $[0, 1]$ - chance for executing TOPSIS mutation (pulling and/or pushing),
    \item t\_best: float $[0, 1]$ - strength of pull towards the attraction point,
    \item t\_worst: float $[0, 1]$ - strength of repulsion from the repulsion point,
    \item best\_individuals\_count: int $[1, population\_size]$ - used to calculate the point of attraction,
    \item worst\_individuals\_count: int $[1, population\_size]$ - used to calculate the point of repulsion.
\end{itemize}
Algorithm characteristics:
\begin{itemize}
    \item Keeping \emph{best\_individuals\_count} best individuals from the population and \emph{worst\_individuals\_count} worst individuals from the population to calculate the point of attraction/repulsion.
    \item Points of attraction and repulsion can be calculated in different ways (multiple variations).
    \item Pulling to the point of attraction with the strength of \emph{t\_best} (where value of 1 would mean to transport the individual straight to this point).
    \item Analogically as above with repulsion point.
    \item This gravity effect is implemented as a mutation and is run on roughly \emph{p * population\_size} individuals.
\end{itemize}

\section{Experimental results}
In order to check the efficacy and efficiency of the proposed algorithms we have conducted a series of experiments using Python 3.10.5, JMetalPy 1.5.5, and Matplotlib 3.5.3 for visualisation purposes.
The experiments were conducted on a PC: Windows 10 Education, Intel Core i5 2500k, 4.2Ghz, 8 GB 1333Mhz.

Frequencies with corresponding observed physical values used to fit the model are shown in Tab.~\ref{tab:data}. 
\begin{table}
    \centering
    \caption{Observation data}
    $$
    \begin{array}{r|l|r|r}
         i & \omega_i & A_i & B_i \\
         \hline
1 & 0.0002 & 0.03238 & -0.00284\\
2 & 0.0003 & 0.03213 & -0.00424\\
3 & 0.0005 & 0.03137 & -0.00694\\
4 & 0.0008 & 0.02962 & -0.01063\\
5 & 0.001 & 0.02813 & -0.01278\\
6 & 0.0012 & 0.02645 & -0.01465\\
7 & 0.0015 & 0.02371 & -0.01692\\
8 & 0.0018 & 0.02087 & -0.01857\\
9 & 0.002 & 0.01899 & -0.01936\\
10 & 0.003 & 0.01063 & -0.02054\\
11 & 0.005 & 0.00057 & -0.01713\\
12 & 0.008 & -0.00540 & -0.01110\\
13 &  0.01 & -0.00704 & -0.00795\\
14 & 0.011 & -0.00757 & -0.00658\\
15 & 0.012 & -0.00795 & -0.00531\\
16 & 0.014 & -0.00843 & -0.00296\\
17 & 0.016 & -0.00860 & -0.00074\\
18 & 0.018 & -0.00846 & 0.00147\\
19 &  0.02 & -0.00795 & 0.00377\\
20 & 0.025 & -0.00346 & 0.00982\\
    \end{array}
    $$
    \label{tab:data}
\end{table}

Our aim was to compare the proposed socio-cognitive metaheuristic with a classic evolutionary aglorithm \cite{michalewicz} implemented in JMetalPy. Moreover, we wanted also to compare the results of our optimized model with the physical one.  All experiments were run 10 times for 15000 evaluations and averages were calculated. 

The algorithms used in the experiments had the following parameters.
Genetic algorithm's parameters:
\begin{itemize}
    \item Population size: 100,
    \item Offspring population size: 20,
    \item Mutation: PolynomialMutation(1.0 / 8.0, 20.0),
    \item Crossover: SBXCrossover(0.9, 20.0),
    \item Selection: BinaryTournamentSelection.
\end{itemize}
All these operators are available in JMetalPy (genetic algorithm).

Caste-based algorithm's parameters:
\begin{itemize}
    \item number\_of\_castes: 3,
    \item chance\_for\_non\_caste\_parents: 0.05,
    \item remaining parameters the same as in evolutionary algorithm.
\end{itemize}

Separated-caste algorithm's  parameters:
\begin{itemize}
    \item number\_of\_castes: 5,
    \item assign\_castes\_interval: 3000,
    \item learn\_from\_better\_caste\_probability: 0.1,
    \item learn\_from\_variable: 0.1,
    \item remaining parameters the same as in evolutionary algorithm. 
\end{itemize}

TOPSIS algorithm's  parameters:
\begin{itemize}
    \item p: 0.1,
    \item t\_best: 0.1,
    \item t\_worst: 0.0,
    \item best\_individuals\_count: 10,
    \item worst\_individuals\_count: 10,
    \item remaining parameters the same as in Genetic
\end{itemize}

\begin{figure*} 
     \centering
     \begin{subfigure}[b]{0.49\textwidth}
         \centering
         \includegraphics*[width=\columnwidth]{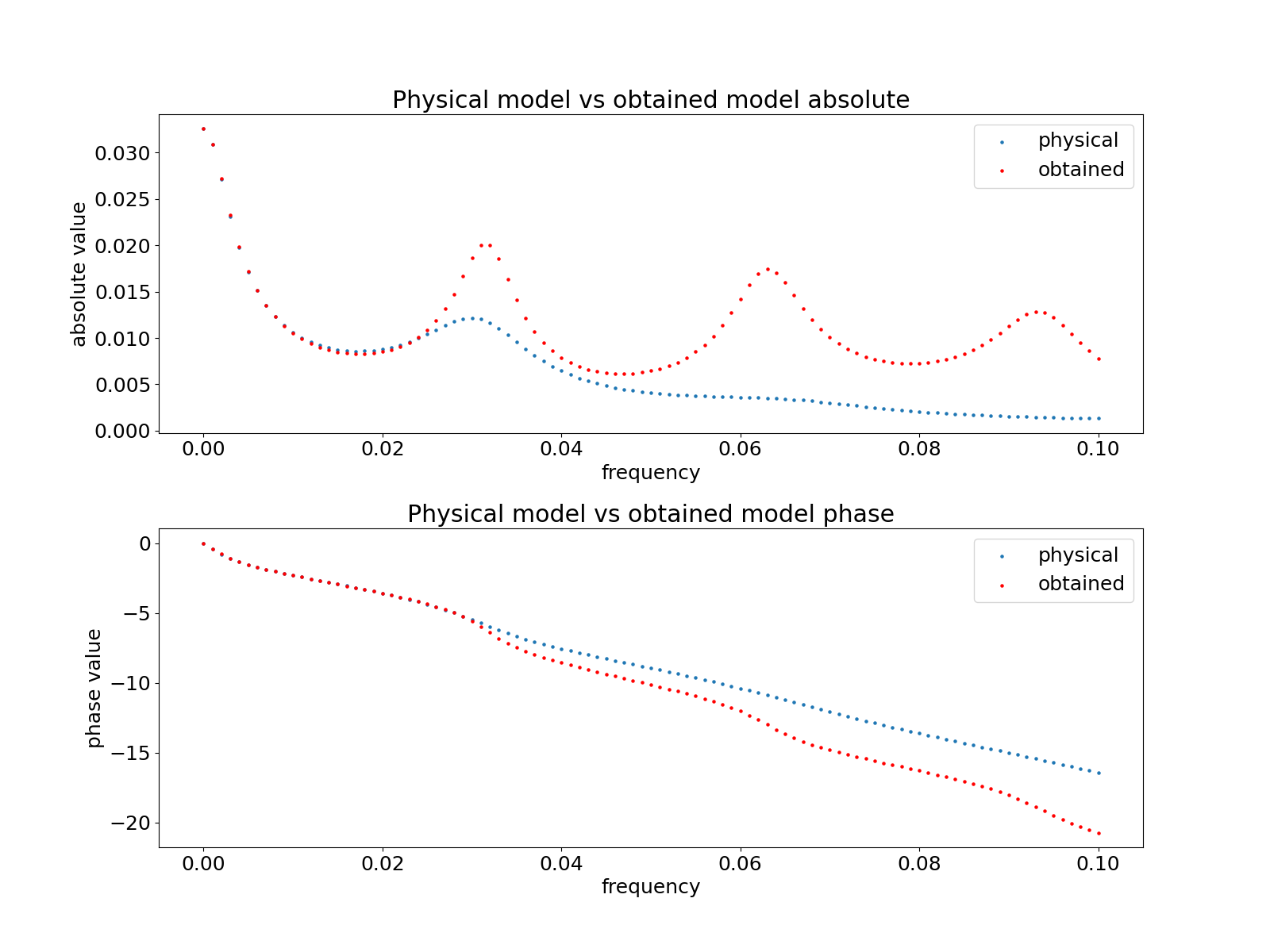}
        \caption{Bode plot for classic genetic algorithm}
        \label{fig:genetic_bode}
     \end{subfigure}
     \hfill
     \begin{subfigure}[b]{0.49\textwidth}
         \centering
         \includegraphics*[width=\columnwidth]{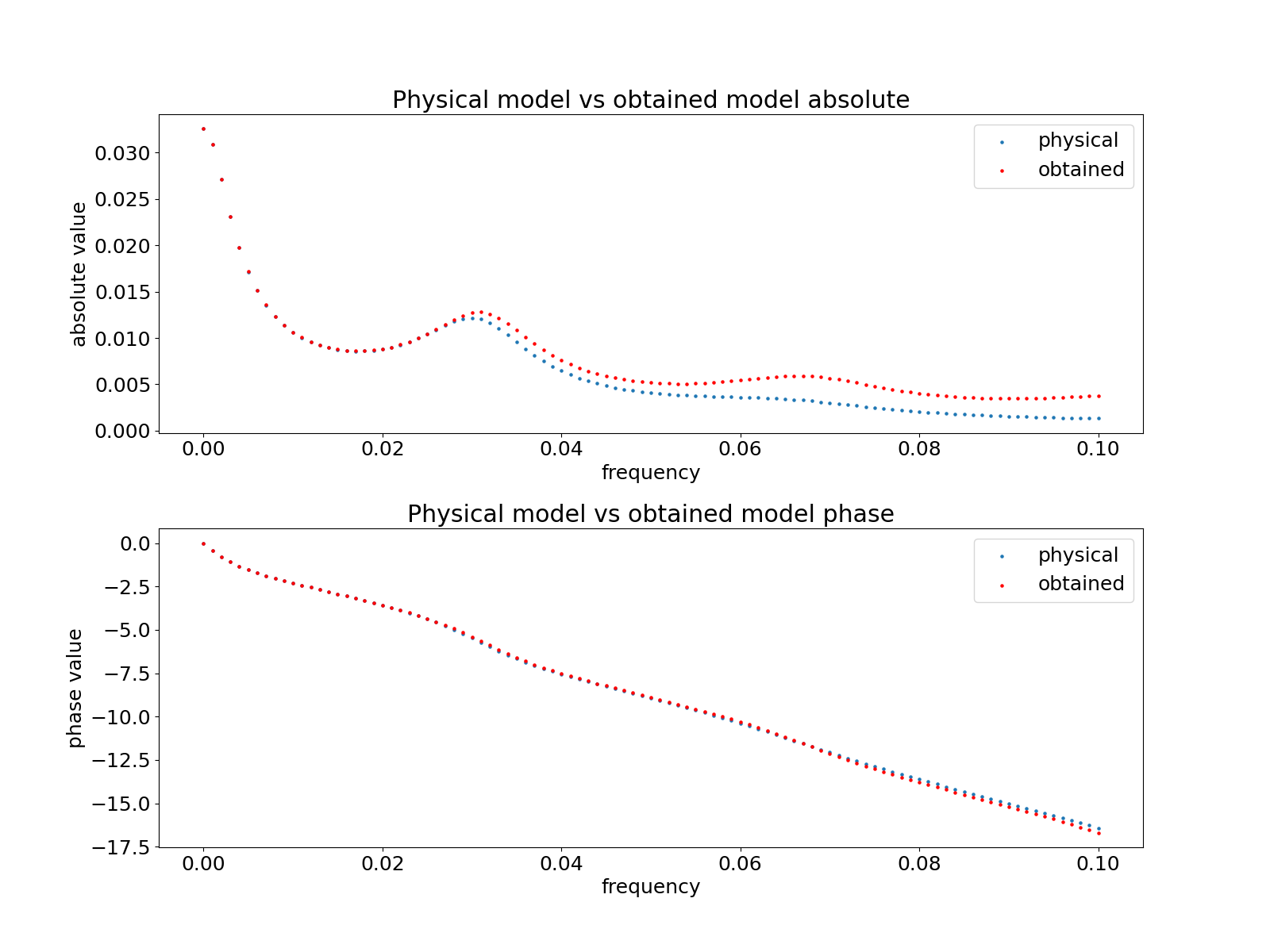}
        \caption{Bode plot for caste-based algorithm}
    \label{fig:caste_bode}
     \end{subfigure}
     \hfill
     \begin{subfigure}[b]{0.49\textwidth}
         \centering
         \includegraphics*[width=\columnwidth]{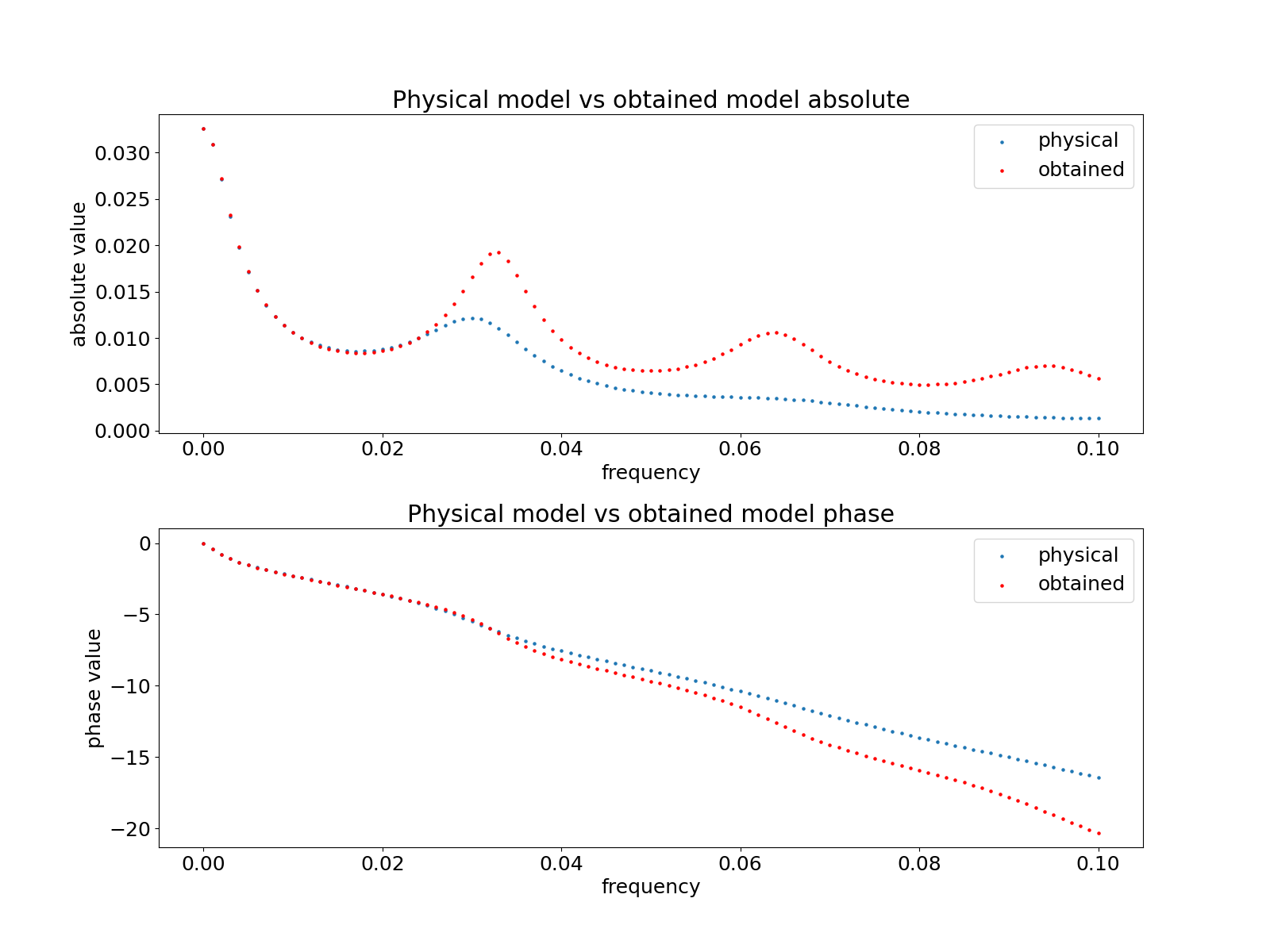}
        \caption{Bode plot for separated caste based algorithm}
        \label{fig:separated_bode}
     \end{subfigure}
     \hfill
     \begin{subfigure}[b]{0.49\textwidth}
         \centering
         \includegraphics*[width=\columnwidth]{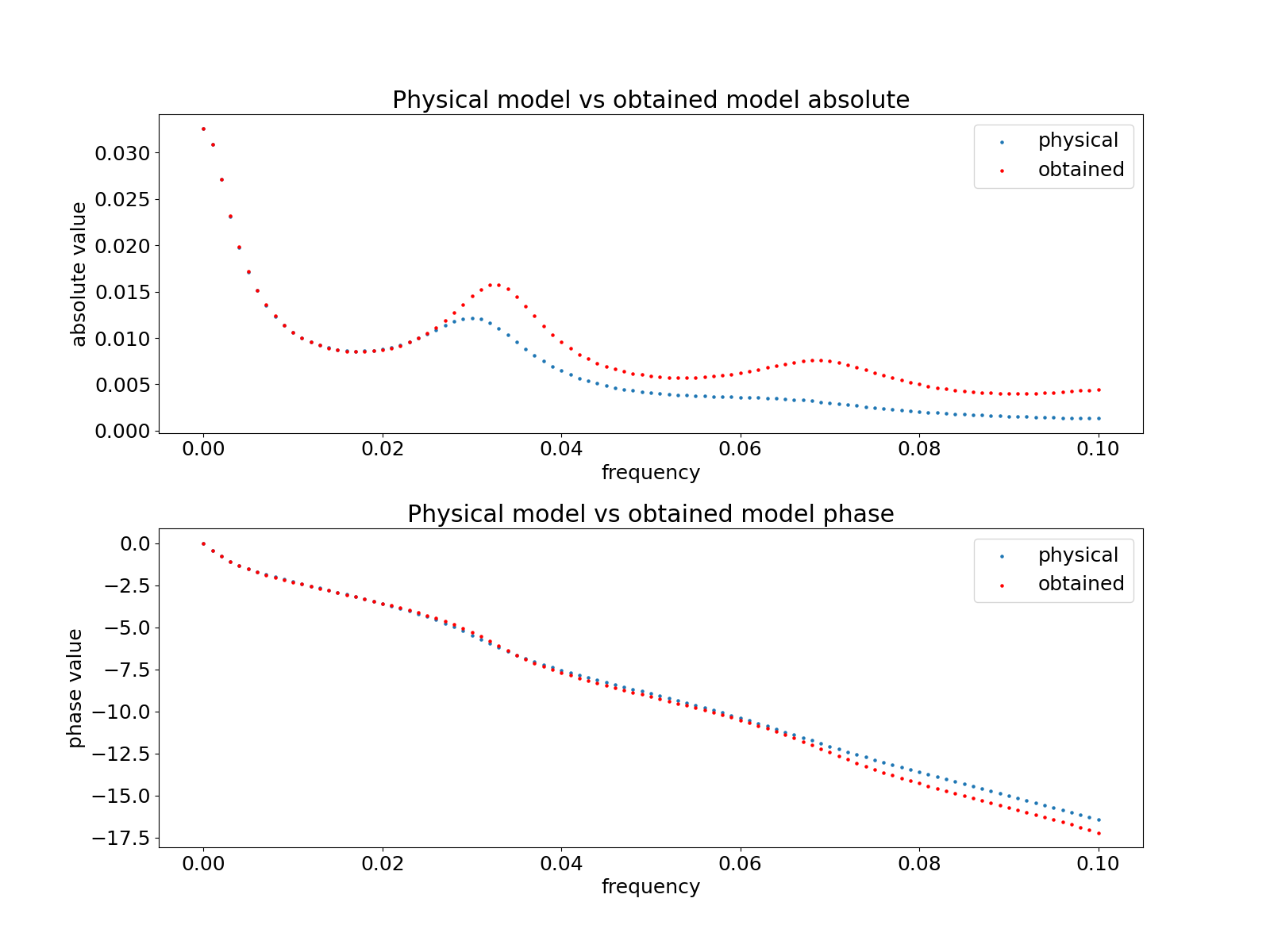}
        \caption{Bode plot for TOPSIS algorithm}
    \label{fig:topsis_bode}
     \end{subfigure}
        \caption{Bode plots for genetic algorithm, caste-based algorithm, algorithms with separated castes and TOPSIS -- comparing the optimized models with the physical one.}
        \label{fig:bodeplots}
\end{figure*}

In Fig. \ref{fig:bodeplots} Bode plots \cite{signals}, showing the  dependency of the amplitude and phase with the frequency  were shown, comparing the optimized models obtained with different algorithms, with the physical ones.
It is easy to see that the model evolved by the classic evolutionary algorithm (using the same parameters for optimization as for the proposed algorithms) is significantly farther from the physical model than the socio-cognitive versions. Comparing the socio-cognitive versions, the Bode plots of the caste-based algorithm seems to be the best fit. The separated caste algorithm is the worst, perhaps the learning relations should be further researched.

\begin{figure*}[p]
     \centering
     \begin{subfigure}[b]{0.49\textwidth}
         \centering
         \includegraphics*[width=\columnwidth]{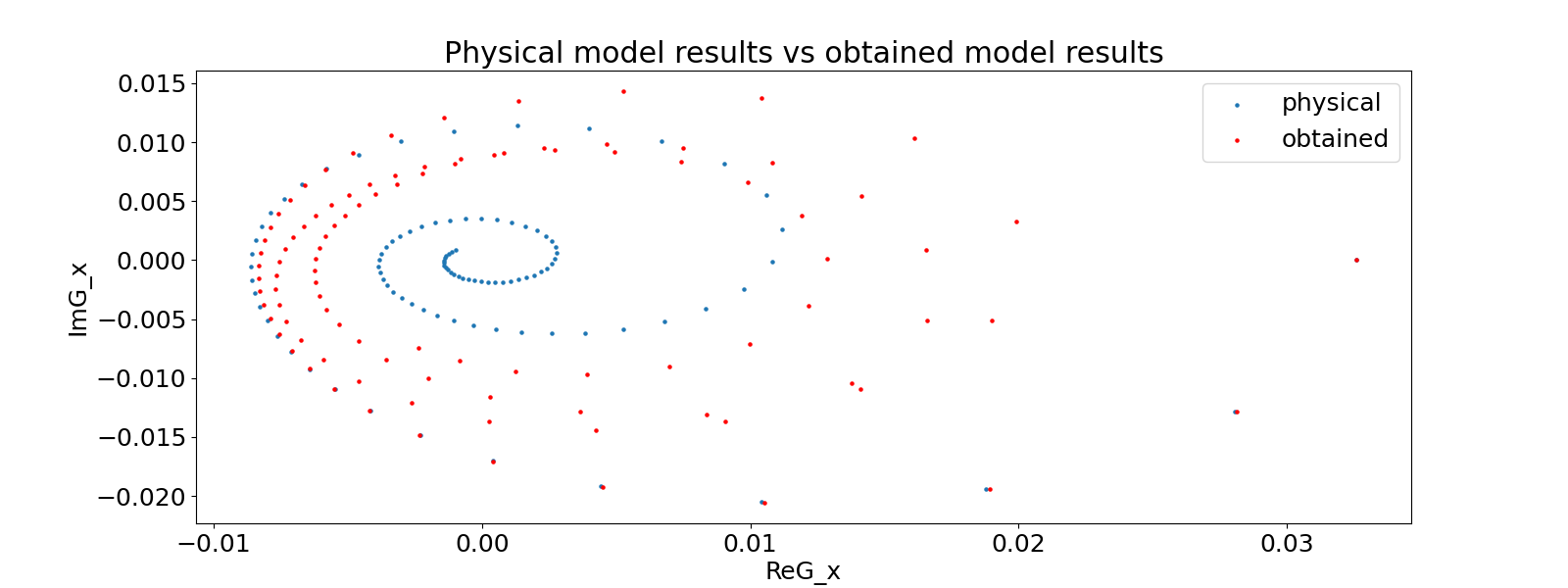}
        \caption{Nyquist plot for classic evolutionary algorithm}
        \label{fig:genetic_nyquist}
     \end{subfigure}
     \hfill
     \begin{subfigure}[b]{0.49\textwidth}
         \centering
         \includegraphics*[width=\columnwidth]{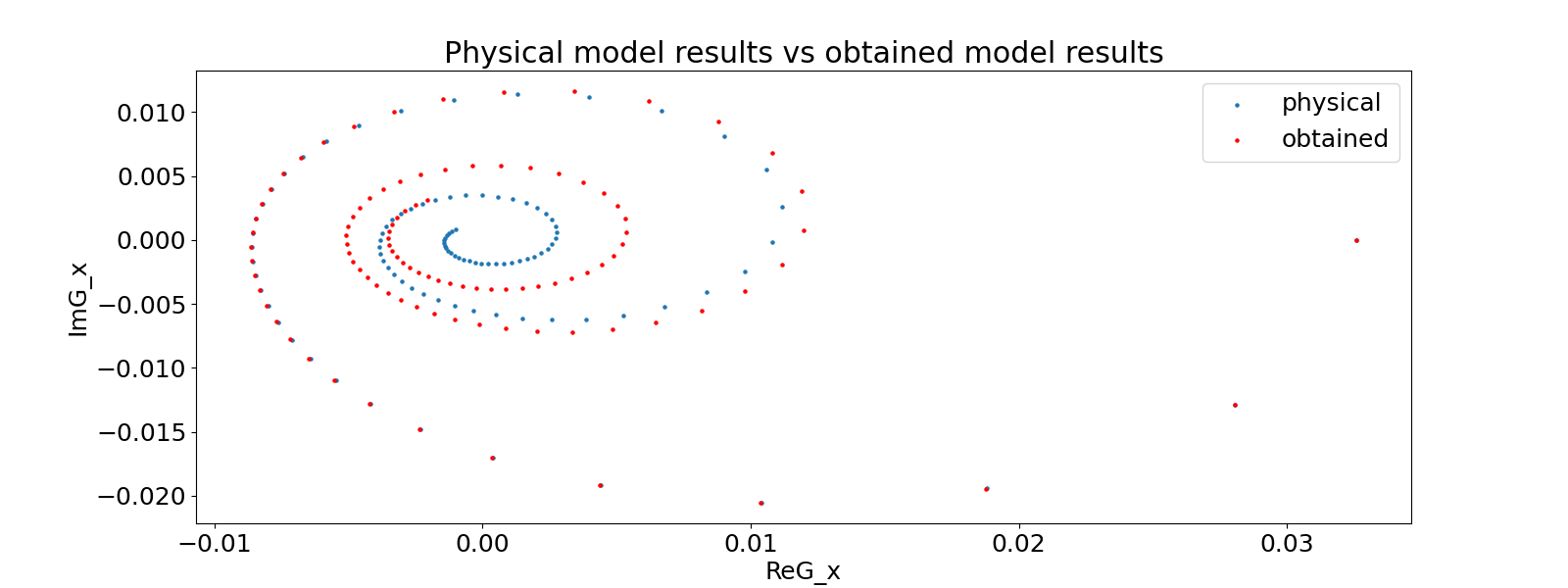}
        \caption{Nyquist plot for caste-based algorithm}
    \label{fig:caste_nyquist}
     \end{subfigure}
     \hfill
     \begin{subfigure}[b]{0.49\textwidth}
         \centering
         \includegraphics*[width=\columnwidth]{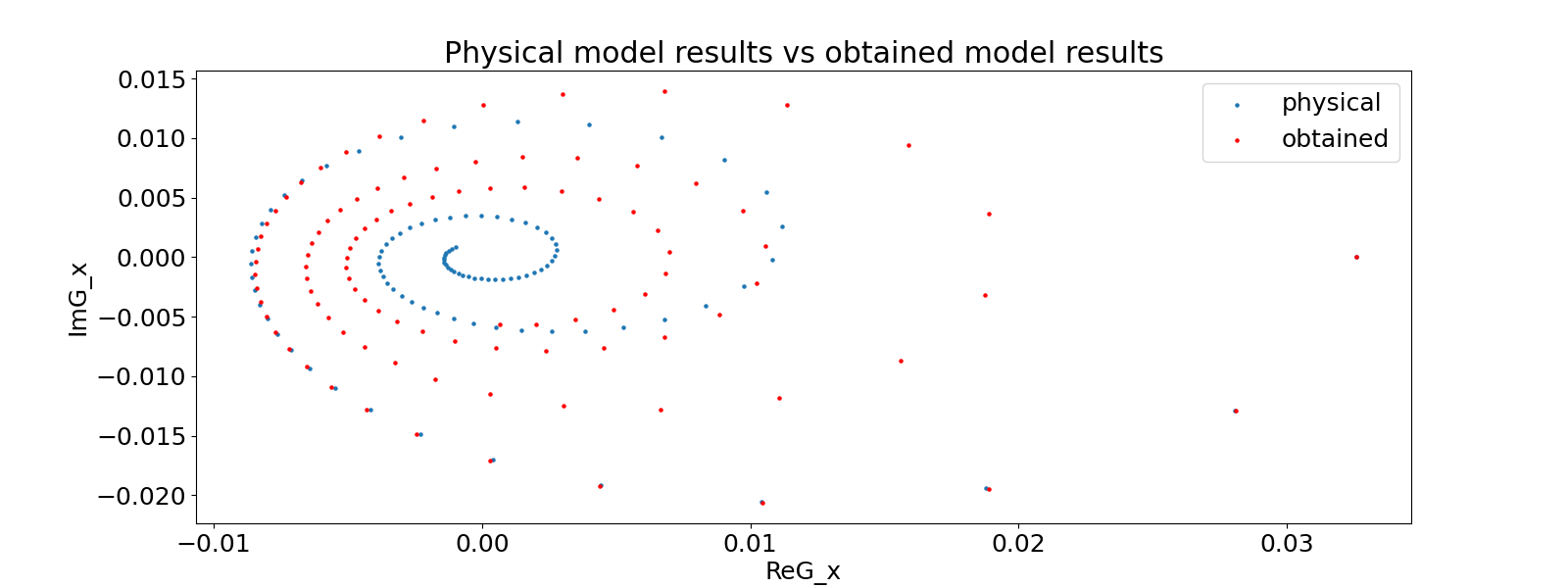}
        \caption{Nyquist plot for separated caste based algorithm}
        \label{fig:separated_nyquist}
     \end{subfigure}
     \hfill
     \begin{subfigure}[b]{0.49\textwidth}
         \centering
         \includegraphics*[width=\columnwidth]{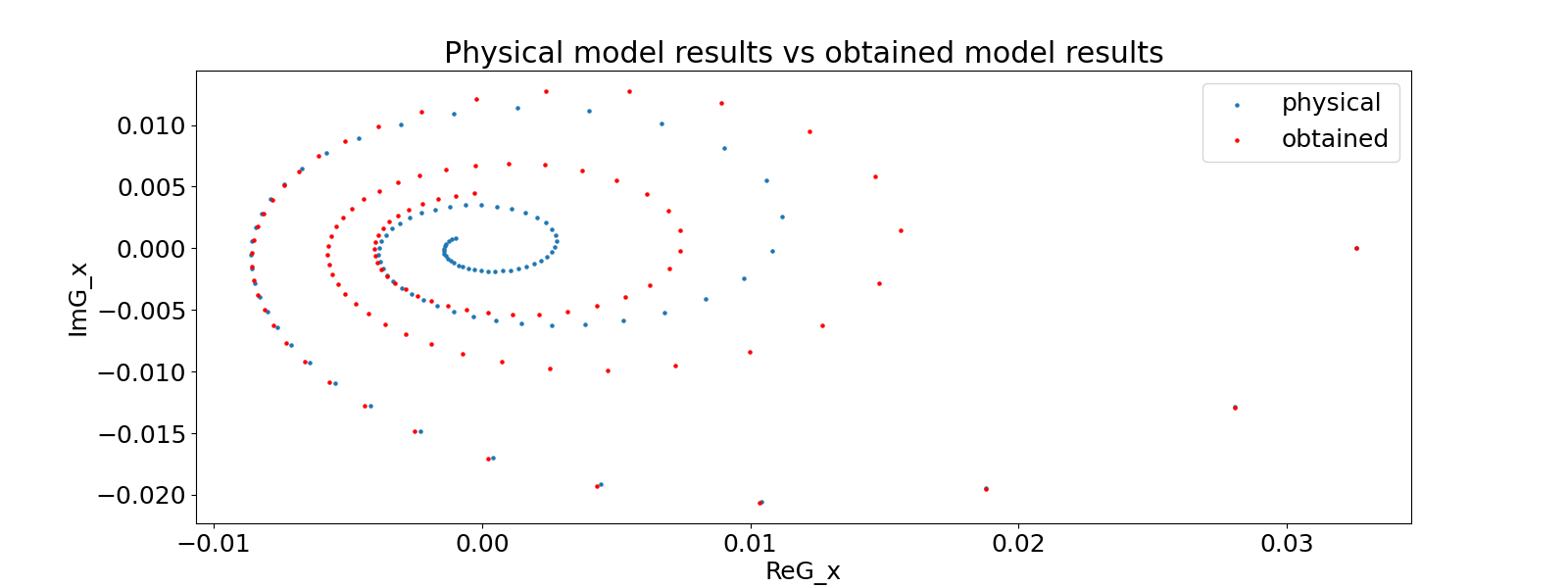}
        \caption{Bode plot for TOPSIS algorithm}
    \label{fig:topsis_nyquist}
     \end{subfigure}
        \caption{Nyquist plots for genetic algorithm, caste-based algorithm, algorithms with separated castes and TOPSIS -- comparing the optimized models with the physical one.}
        \label{fig:nyquistplots}
\end{figure*}

In Fig. \ref{fig:nyquistplots} we can observe the so-called Nyquist plots which are used 
 for assessing the stability of a system with feedback \cite{nyquist}. For our interpretation
 the most important observation is the fitting of the optimized models with the physical ones. Similarly to the above-shown Bode plots (Fig. \ref{fig:bodeplots}) the best fit models is the one produced by the caste-based algorithm. It confirms again, that this variant has the highest potential for application, while the other two socio-cognitive ones (separated caste-based and TOPSIS-based are to be further researched), even though they are also better fit than the one obtained by the classic evolutionary algorithm.



\begin{figure*}[p]
     \centering
     \begin{subfigure}[b]{0.49\textwidth}
         \centering
         \includegraphics*[width=\columnwidth]{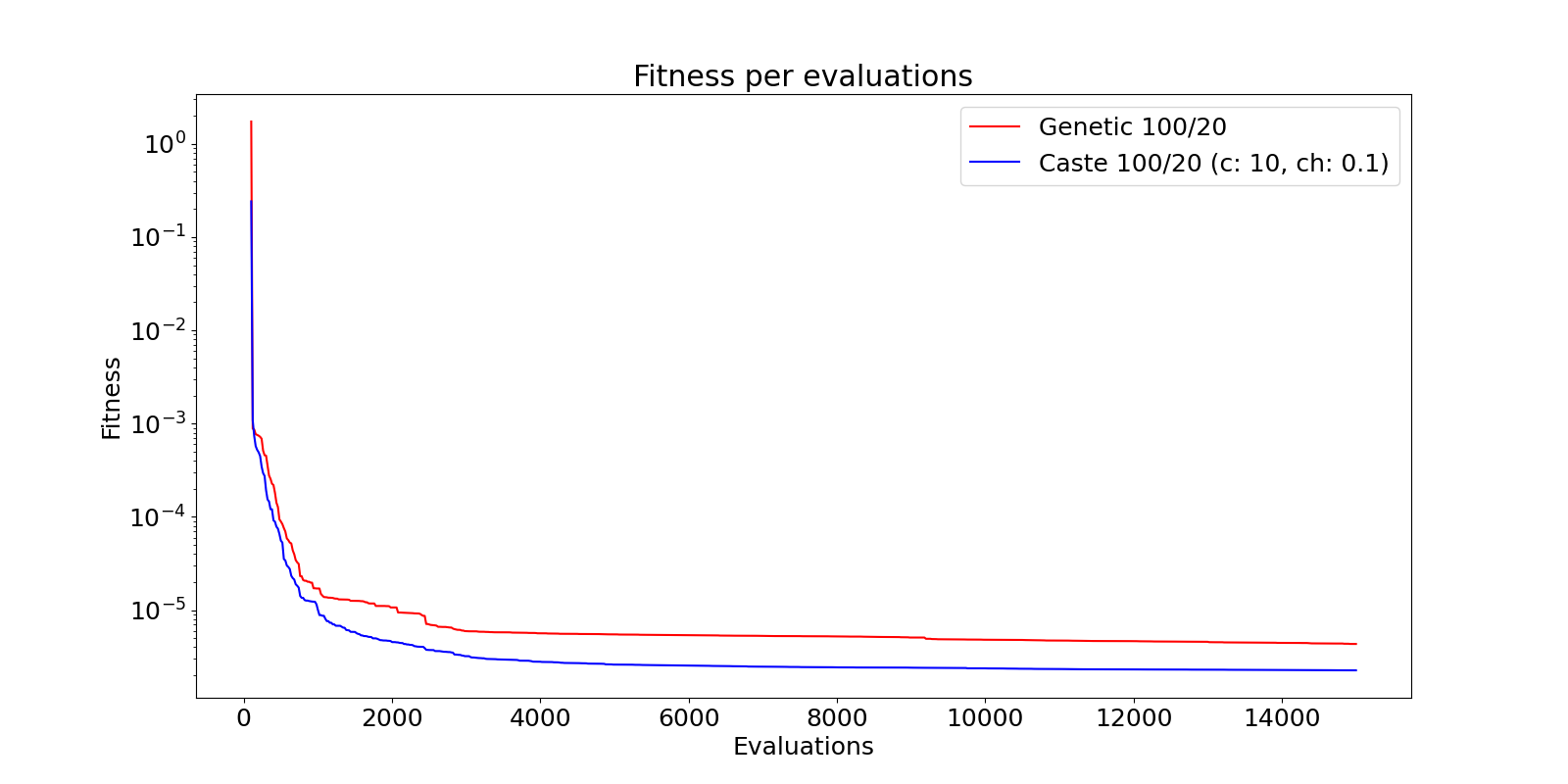}
        \caption{Comparison of caste-based algorithm with genetic one.}
     \end{subfigure}
     \hfill
     \begin{subfigure}[b]{0.49\textwidth}
         \centering
         \includegraphics*[width=\columnwidth]{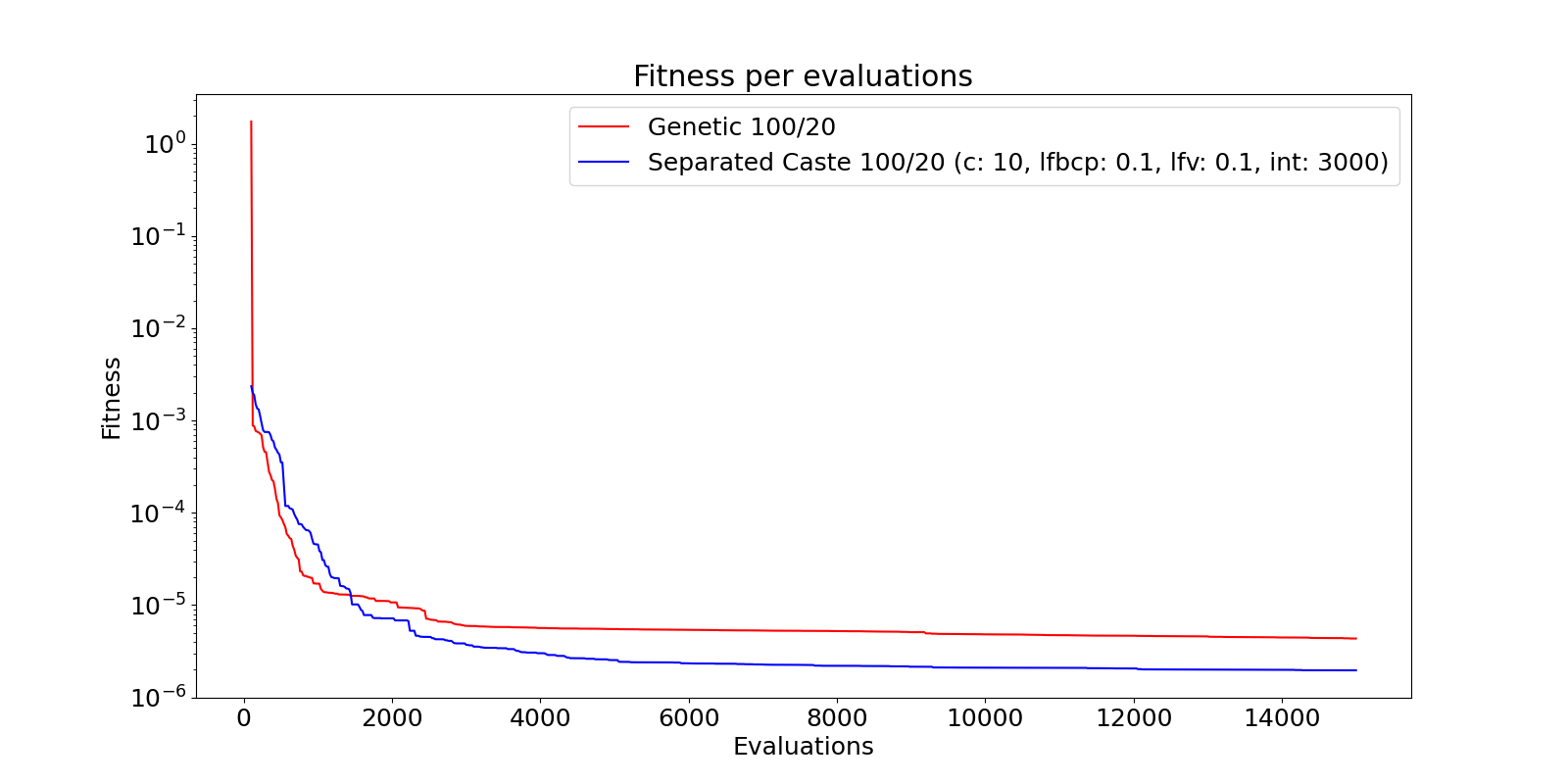}
        \caption{Comparison of separated-caste based algorithm with genetic one.}
     \end{subfigure}
     \hfill
     \begin{subfigure}[b]{0.49\textwidth}
         \centering
         \includegraphics*[width=\columnwidth]{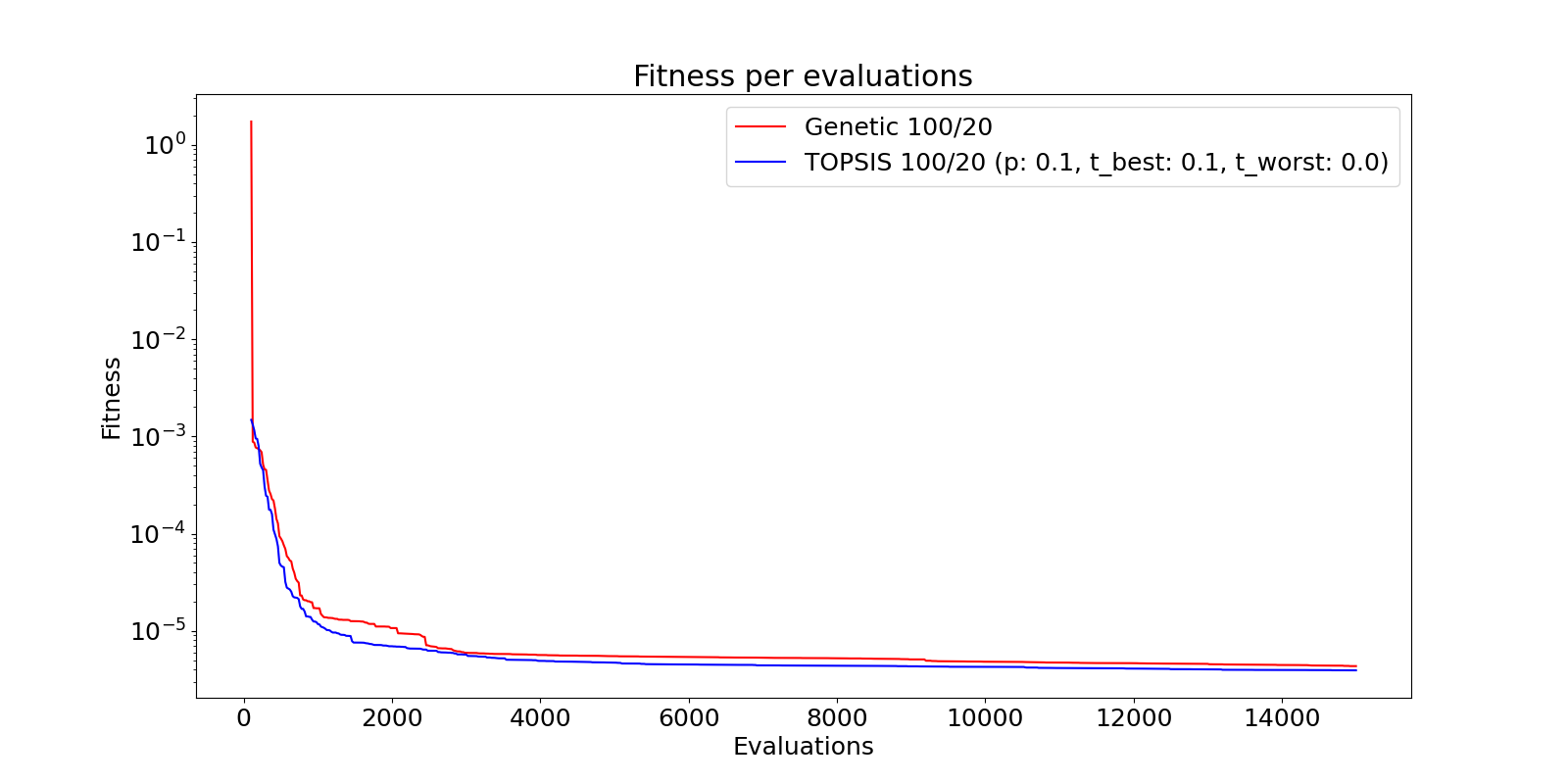}
        \caption{Comparison of TOPSIS-based  algorithm  with genetic one.}
     \end{subfigure}
        \caption{Comparison of dependency between fitness
        and number of evaluation for evolutionary  algorithm, caste-based algorithm, algorithms with separated castes and TOPSIS.}
        \label{fig:comparison}
\end{figure*}

In Fig. \ref{fig:comparison} we can observe the dependency of the best fitness of  the subsequent evaluation of the fitness function. In Y-axis the logarithmic scale was used. It is apparent that all the proposed metaheuristics turned out to be better in the observed cases, although 
they do not completely dominate the classic evolutionary algorithm.

\begin{figure}[h]
   \centering
   \includegraphics*[width=\columnwidth]{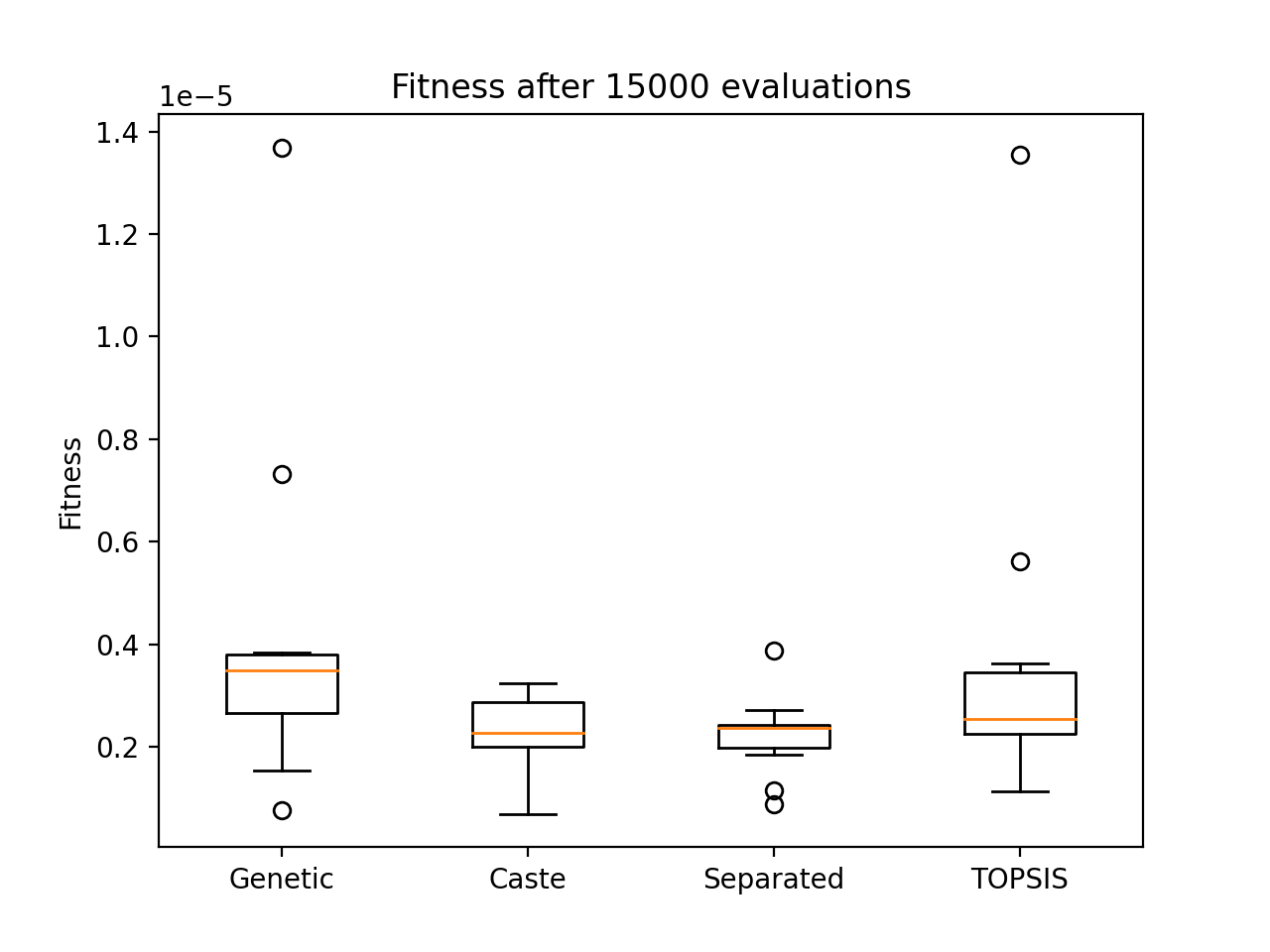}
   \caption{Fitness comparison of all algorithms after 15000 evaluations}
   \label{fig:box_comparison}
\end{figure}

In Fig. \ref{fig:box_comparison} we can see fitness comparison between all algorithms. Moreover, those visualized results may be checked in Table \ref{table:fitnessTable}. It is easy to see that all the tested novel algorithms produced rather comparable results, visibly better than the classic evolutionary algorithm. If we consider minimum fitness obtained in all the experiments, the caste-based algorithm wins, if we think about the repeatability of the results, the most consistent ones are produced for the separated-caste algorithm. The improvement of the novel proposed algorithms is apparent, though they need some more work in order to test them in a broader sense,  using other benchmarks and variants of parameters.
\begin{table}[h]
\centering
\begin{tabular}{| c | c | c | c |}
    \hline
    \textbf{Algorithm} & \textbf{Average} & \textbf{Minimum} & \textbf{Std}\\
    \hline
    Genetic & 4.34e-06 & 7.72e-07 & 3.52e-06\\
    \hline
    Caste & 2.26e-06 & 6.98e-07 & 7.48e-07\\
    \hline
    Separated & 2.24e-06 & 8.78e-07 & 7.88e-07\\
    \hline
    TOPSIS & 3.82e-06 & 1.14e-06 & 3.45e-06\\
    \hline    
\end{tabular}
\caption{Fitness values after 15000 evaluations}
\label{table:fitnessTable}
\end{table}
\section{Conclusion}
In this paper we have shown the possibilities of extension classic evolutionary algorithms by introducing  socio-cognitive inspirations, namely caste-based algorithm into evolutionary one (Michalewicz type). The presented results may be treated as preliminary ones, although they already show the improvement of the algorithms applied to optimization of a very specific problem, namely looking for the best parameters of time-delay system, an important model from the area of automatics.

We have also applied TOPSIS method in order to improve the knowledge transmission between the entities which are to ``learn'' about the others. Moreover, the caste-based algorithm was presented in two variants, with separated and non-separated castes. 

In future we aim at realization of broader experimental results employing more variants of the considered problems and also working-out novel variants of the proposed socio-cognitive metaheuristics.

\section*{Acknowledgments}
The research presented in this paper received partial funding from the Polish National Science Center Project No. 2020/39/I/ST7/02285 and from the funds assigned to AGH University of Science and Technology by the Polish Ministry of Science and Education.

\bibliographystyle{IEEEtran}
\bibliography{bibliography}

\end{document}